\documentclass{article}

\usepackage{microtype}
\usepackage{graphicx}
\usepackage{caption}
\usepackage[justification=centering]{subcaption}
\usepackage{booktabs} % for professional tables

\usepackage{multirow}
\usepackage[most]{tcolorbox}

\usepackage{xurl}
\usepackage{booktabs}
\usepackage{float}
\usepackage{tabularx}
\usepackage{hyperref}

\usepackage[accepted]{icml2024}

\usepackage{amsmath}
\usepackage{amssymb}
\usepackage{mathtools}
\usepackage{amsthm}

\usepackage[capitalize,noabbrev]{cleveref}

\theoremstyle{plain}

\theoremstyle{definition}

\theoremstyle{remark}

\usepackage[textsize=tiny]{todonotes}

\icmltitlerunning{Will we run out of data? Limits of LLM scaling based on human-generated data}

\begin{document}

\twocolumn[
\icmltitle{Will we run out of data? Limits of LLM scaling based on human-generated data}

% It is OKAY to include author information, even for blind
% submissions: the style file will automatically remove it for you
% unless you've provided the [accepted] option to the icml2024
% package.

% List of affiliations: The first argument should be a (short)
% identifier you will use later to specify author affiliations
% Academic affiliations should list Department, University, City, Region, Country
% Industry affiliations should list Company, City, Region, Country

% You can specify symbols, otherwise they are numbered in order.
% Ideally, you should not use this facility. Affiliations will be numbered
% in order of appearance and this is the preferred way.
\icmlsetsymbol{equal}{*}

\begin{icmlauthorlist}
\icmlauthor{Pablo Villalobos}{epoch}\icmlauthor{Anson Ho}{epoch}
\icmlauthor{Jaime Sevilla}{epoch,ab}
\icmlauthor{Tamay Besiroglu}{epoch,mit}
\icmlauthor{Lennart Heim}{epoch,govai}
\icmlauthor{Marius Hobbhahn}{epoch,tu}
\end{icmlauthorlist}

\icmlaffiliation{epoch}{Epoch}
\icmlaffiliation{ab}{University of Aberdeen}
\icmlaffiliation{mit}{MIT CSAIL}
\icmlaffiliation{govai}{Centre for the Governance of AI}
\icmlaffiliation{tu}{University of
Tübingen}

\icmlcorrespondingauthor{Pablo Villalobos}{pablo@epochai.org}

% You may provide any keywords that you
% find helpful for describing your paper; these are used to populate
% the "keywords" metadata in the PDF but will not be shown in the document
\icmlkeywords{Machine Learning, ICML}

\vskip 0.3in
]

\printAffiliationsAndNotice

\begin{abstract}
%Recent progress in language modeling has relied heavily on the availability of human-generated public text data for training large language models. However, as LLMs continue to scale, a question arises: could the availability of data limit their development?
We investigate the potential constraints on LLM scaling posed by the availability of public human-generated text data. We forecast the growing demand for training data based on current trends and estimate the total stock of public human text data. Our findings indicate that if current LLM development trends continue, models will be trained on datasets roughly equal in size to the available stock of public human text data between 2026 and 2032, or slightly earlier if models are overtrained. We explore how progress in language modeling can continue when human-generated text datasets cannot be scaled any further. We argue that synthetic data generation, transfer learning from data-rich domains, and data efficiency improvements might support further progress.
\end{abstract}

\begin{figure}[h]
\includegraphics[width=0.5\textwidth]{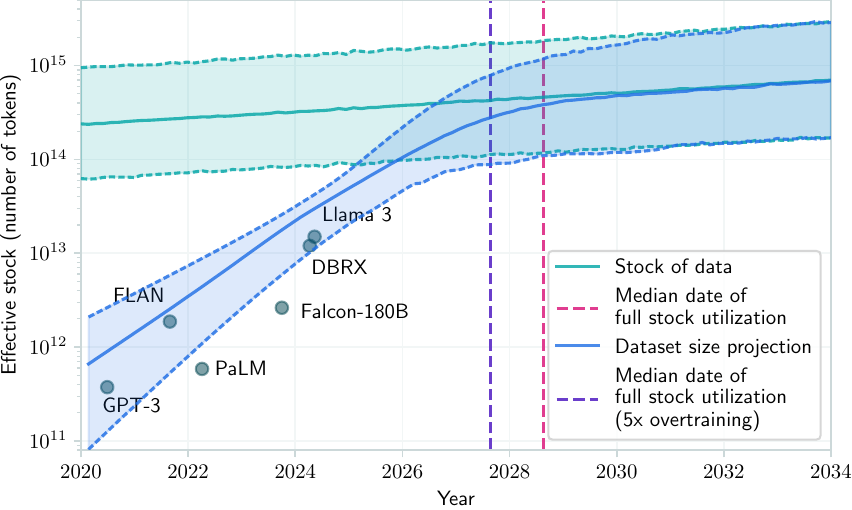}
\caption{\small Projections of the effective stock of human-generated public text and dataset sizes used to train notable LLMs. The intersection of the stock and dataset size projection lines indicates the median year (2028) in which the stock is expected to be fully utilized if current LLM development trends continue. At this point, models will be trained on dataset sizes approaching the total effective stock of text in the indexed web: around 4e14 tokens, corresponding to training compute of $\sim$5e28 FLOP for non-overtrained models. Individual dots represent dataset sizes of specific notable models. The model is explained in Section \ref{sec:exhausting}}
\label{fig:flagship}
\end{figure}

\section{Introduction}
\label{sec:introduction}

Recent progress in language modeling has relied heavily on unsupervised training on vast amounts of human-generated text, primarily sourced from the web or curated corpora \cite{zhao2023survey}. The largest datasets of \textit{human-generated public text data}, such as RefinedWeb, C4, and RedPajama, contain tens of trillions of words collected from billions of web pages \cite{penedo2023refinedweb, redpajama}.

%\newpage

The demand for public human text data is likely to continue growing. In order to scale the size of models and training runs efficiently, large language models (LLMs) are typically trained according to neural scaling laws \cite{kaplan2020, hoffmann2022}. These relationships imply that increasing the size of training datasets is crucial for efficiently improving the performance of LLMs.

\newpage

In this paper, we argue that human-generated public text data cannot sustain scaling beyond this decade. To support this conclusion, we develop a model of the growing demand for training data and the production of public human text data. We use this model to predict when the trajectory of LLM development will fully exhaust the available stock of public human text data. We then explore a range of potential strategies to circumvent this constraint, such as synthetic data generation, transfer learning from data-rich domains, and the use of non-public data.\footnote{The code used in our analysis can be found at \href{https://epochai.org/code/data-stock}{https://epochai.org/code/data-stock}.}

\subsection{Related work}
\label{sec:related-work}

\textbf{Stock of internet data} Several studies have sought to quantify the internet's size and information content. \citet{sizingtheinternet2000} estimated the internet's size at approximately 2.1 billion unique web pages containing 21 terabytes of data. \citet{sizeinternet1998} and \citet{odlyzko2016growth} found that public internet traffic experienced a rapid growth rate of approximately 100\% per year in the early 1990s, which slowed down to double-digits in the late 2010s, particularly in developed countries.

More recently, \citet{idc2018} estimate the total amount of new data created, captured, or replicated worldwide in any given year to be 33 billion terabytes. Unfortunately, the analysis does not break this down into different data modalities (e.g. images, videos, or text data). Focusing just on Google's index, \citet{vandenBosch2016} estimated the stock from 2006 to 2015, finding that it varied significantly over time but is on the order of tens of billions of web pages.

\textbf{Data bottlenecks in machine learning} \citet{muennighoff2023scaling} studied several techniques to mitigate data scarcity for training LLMs. In particular, they considered repeating data, adding more code data, and relaxing the quality filters used during data preprocessing. They quantified the loss of performance when using these techniques to compensate for a smaller data budget, finding that both repeating data and including more code data can compensate for a decrease of up to 75\% in the text data budget. \citet{xue2023repeat} also studied multi-epoch training as a solution for data scarcity. \citet{chinchillaswild2022} argued that high-quality training data would soon become a bottleneck for machine learning.

Leading AI researchers have expressed concerns about data availability limiting the progress of machine learning systems. Dario Amodei, the CEO of Anthropic, estimates a 10\% chance that the scaling of AI systems could stagnate due to insufficient data \cite{roose2023dario}. This underscores the importance of investigating the limitations posed by the finite supply of public human text data.

\begin{table}[ht]
\centering
\begin{tabularx}{\columnwidth}{@{}lXr@{}} % Corrected to four columns; two regular right-aligned
\toprule
Estimate & Median & 95\% CI \\ \midrule
Common Crawl & 130T & [100T, 260T] \\
Indexed web & 510T & [130T, 2100T]\\
Whole web & 3100T & [1900T, 5200T] \\
Images & 300T & N/A \\
Video & 1350T & N/A \\
\bottomrule
\end{tabularx}
\caption{Estimates of the stock of data on the web in tokens.\footnotemark  In the case of images and video we only have point estimates.}
\label{table:stock_estimates}
\end{table}

\footnotetext{Video and image stock estimates are transformed into an equivalent number of text tokens as explained in Appendix \ref{appendix:nontext-data}.}

\section{A model of data scarcity}
\label{sec:exhausting}

The core question we aim to answer is whether the limited availability of public human text data could constrain further LLM scaling. We consider two key variables: the total amount of public human text data available for use (``data stock") and what quantity of this data is actually used in practice during LLM training (``dataset size"). In this section, we develop a model to project both the data stock and dataset sizes.

\subsection{Quantifying dataset sizes}

Specifying our model requires being explicit about how we quantify ``data". To this end, we define the \textit{dataset size} as the number of \emph{tokens} in the training dataset of interest.\footnote{Tokenization is the process of encoding text or other types of data using discrete symbols that can be fed into models \citet{zhao2023survey}. The resulting discrete symbols are known as tokens. The most common choice today is sub-word tokenization, in which each token corresponds to a piece of a word.} In large samples of English text, one token usually corresponds to around 0.8 words (see Appendix \ref{appendix:tokenization}).\footnote{The number of tokens in the dataset should not be confused with the number of tokens seen during training, which could be greater than the dataset size if training occurs over multiple epochs.} 

One limitation of this definition is that the size of a text corpus in tokens depends on how the text is tokenized. That said, in practice, the number of tokens in a corpus does not vary greatly between common tokenizers.\footnote{We make our estimates based on cl100k\_base, a byte-pair encoding (BPE) tokenizer from OpenAI \cite{tiktoken}. In Appendix \ref{appendix:tokenization} we show that commonly used tokenizers produce between 0.5 and 0.2 tokens per byte of text, so we expect our results to be similar using other common tokenizers. In the case of image and video we explain the tokenization estimates in Appendix \ref{appendix:nontext-data}.} Moreover, the two most prominent alternatives -- the number of words and the storage size in bytes -- can vary significantly between modalities or even be undefined.\footnote{For example, the definition of a ``word" can be ambiguous (e.g. in languages which do not use spaces), and in the case of code there are no well-defined words. Meanwhile, storage size can vary by orders of magnitude depending on the choice of compression.}

\subsection{Estimating data stocks}

The first main variable of our model is the data stock $S$. We estimate this by calculating the size of the indexed web and the amount of data that is contained in the average web page, using statistics from Common Crawl.

Since web data contains many low-quality segments of text that do not contribute to model performance \cite{penedo2023refinedweb}, we adjust our estimate to account for differences in data quality. We also adjust for the possibility of multi-epoch training. We explain these adjustments in greater detail in Section \ref{sec:adjustments-data}. As a further robustness check, we estimate the amount of internet text generated each year based on the world population. Table \ref{table:stock_estimates} shows the results of these estimates.

We model our uncertainty about all the observed variables of our models as log-normal distributions, and report our 95\% confidence intervals (CIs) for each of them. The CIs for the latent variables are obtained by Monte Carlo simulations of the functional relationships that define those variables.

\subsubsection{Indexed web}

Common Crawl, a regularly updated open-source collection of scraped web data consisting of over 250 billion web pages \cite{CommonCrawl},\footnote{We use the term ``web page'' to refer to individual pages within a domain or website, for example a single article in Wikipedia.}  serves as the basis for most open web datasets, such as RefinedWeb, C4, and RedPajama. As a subset of the indexed web, Common Crawl's maximum size is inherently bounded by the size of the indexed web.\footnote{Although some web pages might be crawled by Common Crawl but not included in any search engine index, possibly due to being considered very low quality, we expect these web pages to be largely eliminated by quality filters in the data pipelines. Therefore, ignoring them should not significantly impact our results.}

To estimate the size of the indexed web, we use the size of Google's index as a proxy.\footnote{Google is the most widely used search engine globally and receives a significant fraction of all web traffic \cite{similarweb}. Consequently, we expect the size of Google's index to approximate the size of the indexed web within a factor of 2-5.} Applying the methodology proposed by \citet{vandenBosch2016}, we estimate that Google's index contains approximately 250 billion web pages, with a 95\% confidence interval ranging from 100 billion to 1200 billion web pages (see Appendix \ref{appendix:google-index}).

Assuming that Common Crawl is a representative sample of the indexed web,\footnote{This is the stated intention of the Common Crawl team, and since the crawling procedure is quite similar for Common Crawl and search indices, given the size of the archive it seems unlikely to have any significant bias.} we can use it to estimate the average amount of plain text bytes per web page. This number has increased over time, from around 6100 bytes in 2013 to about 8200 bytes in 2021.\footnote{The total size of web pages is 10-20 times larger, since it also includes HTML code, scripts and other non-plain-text data.} We estimate the average plain text bytes per web page to be 7000 [95\%: 6100, 8200].\footnote{We use square brackets to denote 95\% confidence intervals.} 

Each token corresponds to 4 bytes of plain text [95\%: 2, 5] (see Appendix \ref{appendix:tokenization}), so the raw stock of tokens on the indexed web in 2024, calculated according to Equation \ref{eq:stock-common-crawl} is around 510 trillion [95\%: 130T, 2100T].

Since 2013, the plain-text size of the average Common Crawl web page has been growing by between 2\% and 4\% each year. However, estimating the growth rate of the total number of web pages is more challenging due to conflicting evidence. The methodology employed by \citet{vandenBosch2016} suggests that the size of Google's index has remained relatively constant over the past decade, which is a counterintuitive result since new web pages are regularly created. Appendix \ref{appendix:google-index} discusses alternative explanations for this apparent lack of growth in Google's index size.

\begin{tcolorbox}[title=Indexed Web Projected Growth] 
\begin{equation}
    S_{IW}(y) = N_{IW} \times B_P \times T_B \times (1+g)^{y-y_0}
    \label{eq:stock-common-crawl}
\end{equation}

where \(S_{IW}(y)\) is the estimate of the current stock of tokens in the indexed web in a given year $y$, $N_{IW}$ is the number of unique web pages in the indexed web, $B_P$ is the average number of bytes per web page, $T_B$ is the average number of tokens per byte, and $g$ is the estimated rate of growth of the total number of tokens.
\end{tcolorbox}

To better estimate the growth rate of the indexed web, we consider several proxies: global IP traffic, link rot rates, and the growth in the number of internet users. Global IP traffic was increasing by 24\% in 2016 \cite{ciscoIndex}, which can be considered an upper bound on the growth rate of web pages, as the majority of traffic corresponds to consumption rather than creation of text data. Conversely, the number of internet users is growing by approximately 2-4\% per year (Section \ref{sec:internet_pop}), and estimates of the link rot rate range from 2\% to 16\% (Appendix \ref{appendix:google-index}). For Google's index size to remain constant, the link rot rate must be offset by the creation of new web pages or links, suggesting possible growth rates of around 10\%.

However, double-digit growth rates would imply that the average internet user is creating significantly more web pages over time, a trend that appears to be contradicted by some observations, such as the roughly constant rate of tweets per user on Twitter \cite{gdeltprojectVisualizingTwitterx2019s} and similar observations for other platforms such as Wikipedia \cite{wikipediaSize}. Given these considerations, we settle on a confidence interval between 0\% and 10\% a year.\footnote{A doubling of the growth rate from 10\% to 20\% over a 10-year period would result in an 0.4 OOM increase in the data stock. However, this is not large compared to the historical growth rate in data usage of 0.38 OOM per year. Therefore, our conclusions are not highly sensitive to variations in the growth rate within this range.}

\subsubsection{Internet population}
\label{sec:internet_pop}

We consider an alternative model of data stocks that explicitly accounts for the process that generates data. This model relies on the observation that much of the internet's text data is user-generated and stored on platforms such as social media, blogs, and forums. While AI-generated text is becoming more prevalent, we exclude it from this model and discuss it in Section \ref{sec:alt-sources}. In principle, we can estimate the amount of public human-generated text data by considering the number of internet users and the average data produced per user, with growth in data generation primarily driven by the increasing number of internet users.

We model the increase in the number of internet users as coming from two contributors: (1) increases in the human population, and (2) increases in ``internet penetration," i.e. the percentage of the population that uses the internet. For the former, we turn to standard projections by the United Nations \cite{UNpopulation2022}. Since internet penetration has broadly followed an S-curve from $\sim$0\% in 1990 to 50\% in 2016 to over 60\% today \cite{owidtechnologyadoption}, we model this using a sigmoid function, fitting it to the data in \citet{owidtechnologyadoption}.

Finally, the amount of data generated per internet user varies across countries and over time due to differences in culture, demographics, socioeconomic factors, and online services. Quantifying these variations is complex and beyond the scope of this analysis, so we assume that the average data production rate per user remains constant to enable a tractable estimate.

\begin{figure}[ht]
  \small
  \centering \includegraphics[width=0.48\textwidth]{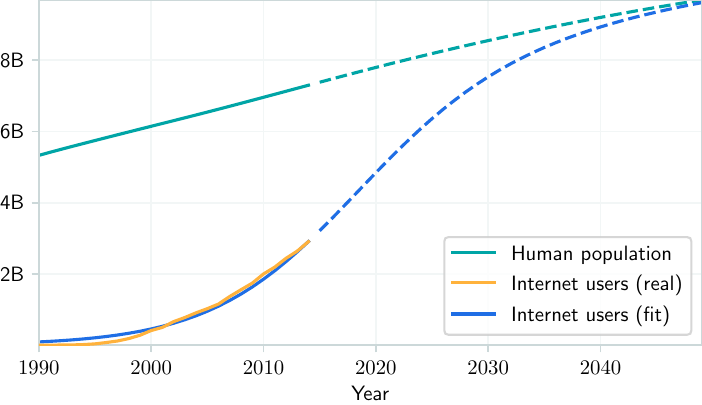}
  \caption{\small Historical and projected evolution of internet users. Historical data is from \citet{owidinternetusers}.}
  \label{fig:internet_pop_model}
\end{figure}

This model of the number of internet users closely matches the historical data (Figure \ref{fig:internet_pop_model}). A more detailed explanation of this model can be found in Appendix \ref{appendix:reddit}.

Based on reported user statistics for major online platforms (see Appendix \ref{appendix:nonpublic-data}), we estimate the total volume of text data uploaded to the internet in 2024 was between 180T and 500T tokens. To project future data accumulation, we scale this initial 2024 estimate by the projected number of internet users in each subsequent year. This provides the estimated annual data contribution from the global online population. We then cumulatively sum these yearly contributions over time to model the total stock of internet text data. The final estimate is 3100T [95\%: 1900T, 5200T] tokens. This estimate includes both data on the indexed web and the deep web, and therefore serves as an upper bound on the size of the indexed web.

\subsection{Data quality and multi-epoch training}
\label{sec:adjustments-data}

The preceding subsections outline the core basis of the model that we use in our analysis. However, before performing forecasts, we first need to account for a few additional considerations. 

\begin{figure}[h]
  \small
  \centering
  \includegraphics[width=0.4\textwidth]{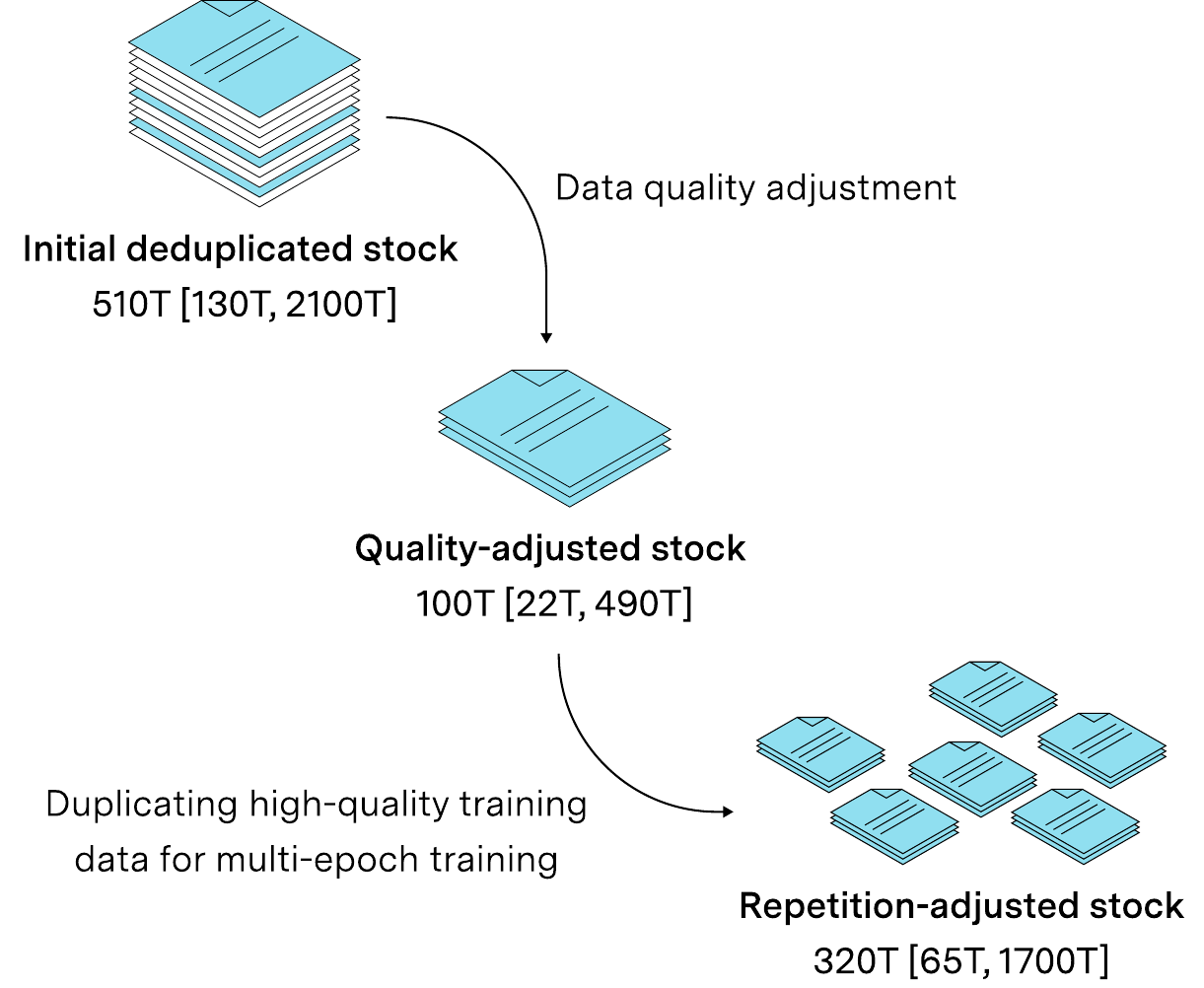}
  \caption{\small Illustration of the adjustments for quality and repetition and the adjusted stock sizes in number of tokens. First the lower-quality data is filtered out, and then the resulting dataset is duplicated for multi-epoch training.}
  \label{fig:stock_model}
\end{figure}

In particular, since our focus is on data constraints in the scaling of \emph{language models}, the literal number of tokens in the training dataset may not be what matters for improving LLM performance. For example, differences in data quality \cite{li2023textbooks} and the number of training epochs \citep{muennighoff2023scaling} can potentially have a substantial effect on final model performance. In this subsection, we analyze the significance of these factors and modify our model accordingly. Our adjustments for data quality and multi-epoch training are illustrated in Figure \ref{fig:stock_model}.

\subsubsection{Data quality}
\label{sec:high_quality_methods}

One way in which only considering the measure of ``number of tokens" is too simplistic is that not all public human text data is created equal. Intuitively, we would expect models that are trained primarily on books or Wikipedia to outperform models that are purely trained on YouTube comments. In this way, public human text data from books are ``higher quality" than YouTube comments. Such intuitions are in fact supported by some empirical observations. For example, data processing techniques like deduplication \cite{lee2022deduplicating} and data filtering \cite{gao2021empirical} have been shown to improve model performance.

However, building in these effects into our model is nontrivial. For one, there is no standard accepted measure of data quality \cite{mitchell2023measuring}. Instead, we are forced to rely on a fairly vague working definition: A dataset is of higher quality than another if training on it leads to higher performance, at similar dataset sizes.% This captures the aforementioned intuition that human-curated datasets, like published papers, books, or news articles, are often considered higher quality than general web data \cite{brown2020language, thePile, scao2022language}. 

Recent findings show that with adequate filtering, data extracted from the web can outperform datasets constructed from human-curated sources \cite{penedo2023refinedweb}. In addition, \citet{xie2023doremi} found that in The Pile, which is a dataset consisting of web data and human-curated sources, increasing the proportion of web data up to 40-70\% led to substantially higher performance. These empirical findings suggest that while much of internet public human text data is on average ``lower quality" than human-curated sources, one can potentially make up for this through careful data processing. 

Given these considerations, we can attempt to determine how much we need to adjust our previous model to account for data quality. We operationalize this in terms of how much ``low quality" data is filtered to achieve optimal performance in practice. \citet{penedo2023refinedweb} create a 5T-token dataset that outperforms curated corpora by carefully filtering and deduplicating raw data from Common Crawl. The filtering part of this process reduced the size of the web dataset by around 30\%. Meanwhile, \citet{marion2023more} found that pruning around 50\% of deduplicated data from a subset of Common Crawl using a perplexity measure led to optimal performance.\footnote{First, the Common Crawl raw data is filtered using regular methods down to 20\% of the original size, and then the remaining data is further pruned using several quality metrics. Figure 4 in \citet{marion2023more} shows that the best result is obtained when pruning 50\% of the deduplicated dataset, for a final size of 10\% of the original. Note that there appears to be no benefit from using the rest of the data, instead of repeating the best 10\%.} Based on these empirical results, we believe with 95\% certainty that between 10\% and 40\% of deduplicated web data can be used for training without significantly compromising performance.

\subsubsection{Multiple epochs}

Besides data quality, using the ``number of tokens" as a measure does not account for the possibility of multi-epoch training. The degree to which stocks should be adjusted for multiple epochs depends on the effectiveness of training on the same data over multiple epochs, compared to training on new ``unique" data. 

\citet{muennighoff2023scaling} investigate this empirically, fitting a scaling law for the performance of a model trained for multiple epochs. Concretely, for a given model trained on multiple epochs, this law gives an estimate of the dataset size that would produce an equally capable model with just one epoch.. This is the ``effective dataset size'' of a multi-epoch training run. The authors estimate the maximum increase in the effective dataset size that can be gained from multiple epochs at between 3x and 15x, and we anchor to this estimate in adjusting our model. Because additional epochs yield diminishing returns, the upper extreme of 15x would require a very inefficient training procedure with a large number of epochs that does not correspond to common practices.\footnote{Typical numbers are between 1 and 4 epochs, for example see \citet{taylor2022galactica} and \citet{touvron2023llama}.} For this reason we reduce it to 5x.

\begin{tcolorbox}[title=Historical Dataset Size Growth Projection] 
\begin{equation}
D_H(y) = G_D^{y - y_0} D(y_0)
\label{eq:historical-dataset-projection}
\end{equation}
where \(D_H\) is the training dataset size, \(G_D\) is the factor growth per year, \(Y_0\) is some base year, and \(Y\) is the year. Both \(G_D\) and \(D(y_0)\) are lognormal distributions.
\end{tcolorbox}

\subsection{Projecting growth in dataset sizes}
\label{sec:projecting_growth}

To project the future values of our second key variable, the training dataset size $D$, we begin by examining historical growth rates and extrapolating them forward.

To estimate historical growth, we use the database of notable  machine learning models in \citet{epoch2023pcdtrends}, a comprehensive database that contains annotations of over 300 machine learning models. We filter this data to include only large language models (LLMs) from papers published between 2010 and 2024, resulting in a subset of around 80 data points. We then perform a linear regression on the logarithm of the dataset size against time, as shown in Equation \ref{eq:historical-dataset-projection}. This yields a median estimate of 0.38 orders of magnitude per year (OOM/y), or around 2.4x per year, with a bootstrapped 95\% confidence interval of 0.27 to 0.48 OOM/y.

\begin{figure}[h]
\includegraphics[width=0.5\textwidth]{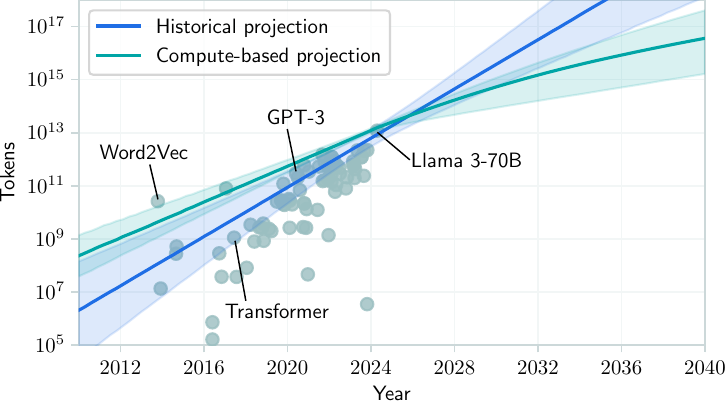}
\caption{\small Projections of data usage. Two extrapolations of data usage, one from past trends and one from compute availability estimations plus scaling laws. The shaded areas denote a 90\% CI for the extrapolated median. The dots are individual training runs.}
  \label{fig:dataset_proj}
\end{figure}

To project this trend forward, we first need to determine the size of the largest datasets used today, which are typically around 10T tokens.\footnote{Skywork-13B \cite{wei2023skywork}, XVERSE-65B \cite{xverse65b}, and PaLM 2 \cite{anil2023palm} were each estimated to be trained on roughly 4T tokens \cite{epoch2023aitrends}. DBRX \cite{databricksIntroducingDBRX} reportedly used 12T tokens, and Llama 3 \cite{llama3} used 15T tokens.}\footnote{Some of the largest recent models, like GPT-4 and Gemini Ultra, do not report the size of their datasets and so we have not included them in the analysis. However, estimates of their training compute are around 5e25 floating point operations (FLOP) \cite{epoch2023aitrends}, which, assuming they are trained using Chinchilla scaling, would correspond to a dataset size of approximately 13T tokens. Therefore, we expect their datasets to be within the same order of magnitude as the biggest ones on our list.} 
Naively projecting the historical trend from this baseline suggests that systems could be trained on over one quadrillion tokens by the end of the decade (see Figure \ref{fig:dataset_proj}).

The historical growth rate in dataset sizes cannot continue indefinitely, even if the data stock was unlimited. In the past, the increasing scale of computing power has driven the demand for larger training datasets, consistent with neural scaling laws for dense transformers which suggest that training data size should scale roughly with the square root of training compute \cite{hoffmann2022, dey2023cerebrasgpt, fetterman2023tune}.

However, the growth in compute is also subject to limits, and current fast rates may not be sustained indefintely. Technical limitations, such as the energy efficiency of computing devices \cite{ho2023limits} and limits on the electricity supply to data centers,\footnote{
In the US, electric grids are already struggling to meet the growing electricity demands from data centers, which are critical infrastructure for AI \cite{Zimmerman2023ReadyToGo}. Upgrading transmission capacity to deliver more power to these facilities involves lengthy planning and construction timelines, often spanning many years (ibid.)} restrict the feasible amount of compute. Other factors like chip production capacity and economic constraints could slow down the rate at which computing power used in training can scale. Consequently, if the ability to scale computing power is constrained, it will likely lead to a deceleration in the historical trends of dataset size growth.

To introduce this constraint into our model, we need estimates of the maximum compute budget for training that will be available in the future. For this purpose, we use the results from \citet{epoch2022projectingcomputetrends}, which performs such a projection based on estimated training compute growth rates in frontier machine learning systems between 2010 and 2022.\footnote{Note that this projection has a wide range of uncertainty and includes scenarios in which spending on compute grows orders of magnitude over current levels, up to 1\% of GWP.} Following \citet{hoffmann2022}, we further assume that compute-optimality involves training on 20 tokens per parameter, per Equation \ref{eq:dataset-size-compute}.

\begin{tcolorbox}[title=Compute-based Dataset Size Growth Projection]
\begin{equation}
    D_C(y) = \sqrt{\frac{20}{6} \cdot C(y)}
    \label{eq:dataset-size-compute}
\end{equation}

where \(D_C(y)\) is the projected amount of data used in notable training runs and \(C(y)\) is the probabilistic projection of largest compute spent on a training run, modeled following \citet{epoch2022projectingcomputetrends}.

$6$ is the number of FLOP per parameter per token and  $20$ is the approximate number of training tokens per parameter according to \citet{hoffmann2022}.
\end{tcolorbox}

As illustrated in Figure \ref{fig:dataset_proj}, the resulting model closely matches the historical trend and its projection until around 2030. It then slows down over time. 

Our final projection of growth in dataset sizes is an equally-weighted mixture of both the historical and compute-based projections \footnote{The historical projection is simpler but seems to contradict reasonable assumptions about compute scaling. A mixture provides a good representation of our uncertainty.} (see Equation \ref{eq:dataset-size-mixture}). It is illustrated in Figure \ref{fig:flagship}.

\begin{tcolorbox}[title=Mixture Projection of Dataset Size Growth]
\begin{equation}
    F_{D(y)} = \frac{1}{2}\left( F_{D_H(y)} + F_{D_C(y)}\right)
    \label{eq:dataset-size-mixture}
\end{equation}

where \(D_H(y)\) is the historical projection of dataset sizes, \(D_C(y)\) is the compute-based projection, and \(F_X\) is the cumulative distribution function of the random variable \(X\).

\end{tcolorbox}

\subsection{When will the stock of public human text data be fully utilized?}

\begin{figure}[ht]
  \small
  \centering
  \includegraphics[width=0.45\textwidth]{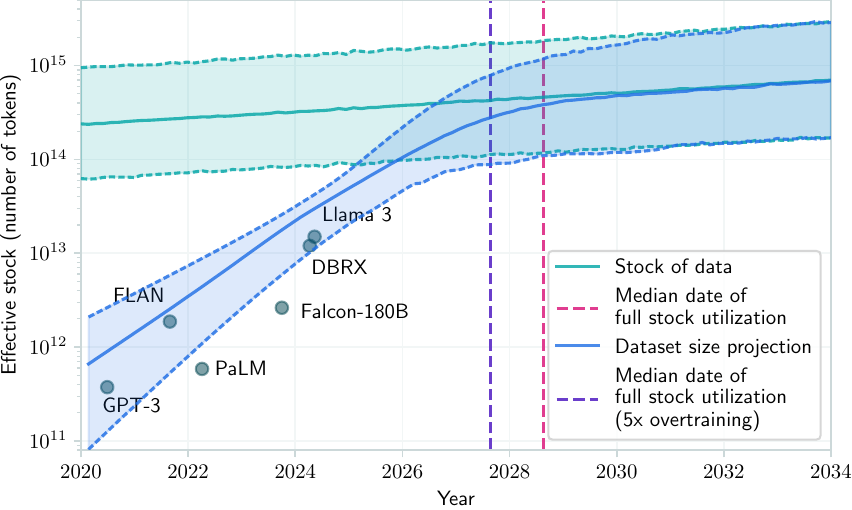}
  \caption{\small Projection of effective stock of human-generated public text and dataset sizes used to train notable LLMs. The intersection of the stock and dataset size projection lines indicates the median year (2028) in which the stock is expected to become fully utilized if current LLM development trends continue. At this point, models will be trained on dataset sizes approaching the total effective stock of text in the indexed web: around 4e14 tokens, corresponding to training compute of $\sim$5e28 FLOP for non-overtrained models.}
  \label{fig:exhaustion_date}
\end{figure}

Combining our projections of dataset size increases, and our estimate of the stock of data, we can estimate when the full stock will be used in a training run if past trends continue. Figure \ref{fig:exhaustion_date} shows the projected availability and usage of effective data. The intersection between these projections corresponds to public text data being exhausted. The median exhaustion year is 2028, and by 2032 exhaustion becomes very likely. At the point the data stock is fully utilized, models will be using around 5e28 FLOP during training.

\begin{figure}[ht]
  \small
  \centering
  \includegraphics[width=0.45\textwidth]{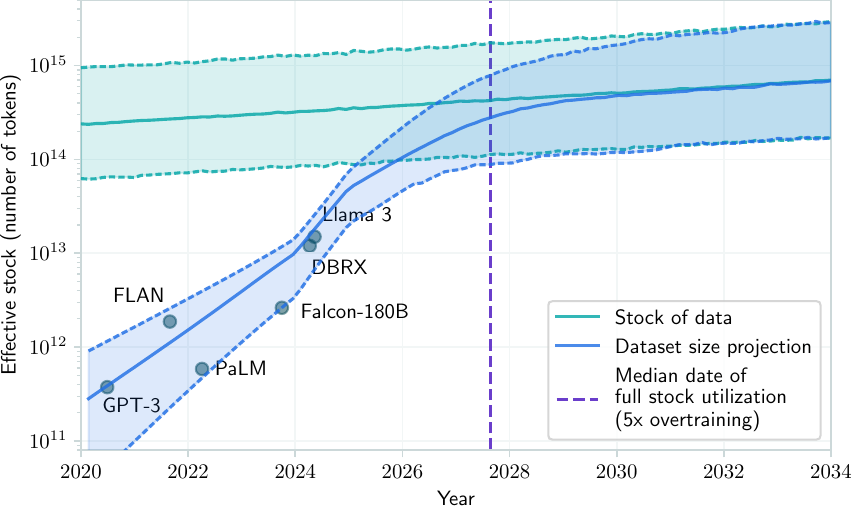}
  \caption{\small Compute-based data usage projection, assuming that frontier models will be overtrained by 5x starting from 2025. This policy results in the stock of data being fully used earlier than with a compute-optimal scaling policy.}
  \label{fig:overtrained_projection}
\end{figure}

An important assumption in our projections is that models are trained compute-optimally\footnote{Compute-optimal training refers to selecting the number of parameters and dataset size of the model to produce the maximum possible capabilities for a given level of training compute \cite{hoffmann2022}.}. However, many developers might instead decide to ``overtrain'' models to achieve better efficiency during inference \cite{sardana2023chinchillaoptimal}, which would require more data. The degree of overtraining that will be chosen by developers depends on a multitude of factors, in particular how many tokens will be generated during inference \cite{sardana2023chinchillaoptimal}, and is hard to predict in advance. That said, based on our analysis in Appendix \ref{appendix:overtraining_model}, we consider overtraining by 5x to be a reasonable choice.\footnote{Overtraining by  5x means that the tokens/parameter ratio is 5x higher than that of a compute-optimal model, or equivalently that the model uses $\sqrt{5}$ times more data than a training-compute-optimal model trained using the same amount of compute.} This would result in a data bottleneck one year earlier than our projections indicate, at a training compute level of $\sim$6e27 FLOP.\footnote{We still decide to focus on compute-optimal training for two reasons. First, we are more interested in frontier capabilities, and at any given compute budget the capabilities of models trained compute-optimally will be higher. Second, normally only smaller models are heavily overtrained. For example, Llama 3 8B is overtrained by close to 100x, while Llama 3 70B is only overtrained by 10x. Models that are closer to the compute frontier tend to be less overtrained due to the high cost of overtraining.}

According to our projections, data could become a significant bottleneck for training LLMs this decade, particularly if LLMs continue to be intensively overtrained. This timeline allows for potentially substantial improvements in LLM performance, given the rapid progress in recent years \cite{ho2024algorithmic, Sevilla_2022}. However, when considering the near 70-year history of AI, this timeframe is relatively short. While significant advancements can be made in the coming years, the impending data bottleneck presents an urgent challenge for the long-term progress of AI. For AI progress to continue into the 2030s, either new sources of data or less data-hungry techniques must be developed. The following sections of this paper will address some of these possibilities.

\section{Beyond public human text data}
\label{sec:alt-sources}

While the core focus of this paper is on public human text data in particular, understanding the broader implications of our model's predictions requires considering ways in which the model might be wrong or incomplete. Crucially, although the model predicts that public human text data will be fully utilized at around the end of the decade, this does \emph{not} necessarily imply that training data will bottleneck ML scaling at that time. In this section, we briefly survey possible ways of circumventing bottlenecks in public human text data.

For example, our model assumes no substantial change in the underlying process of increasing the public human text data stock. One naive way in which this assumption breaks is if significantly more humans are paid to generate more text. While this might be valuable at small scale for certain types of data, it is unlikely to be an economical way to generate an appreciable increase in text for general-purpose pre-training.\footnote{Appreciably increasing the stock of text data could require hiring millions of people. 10 million people writing 40 words per minute for 8 hours per day would write 70T words in one year, which is the same order of magnitude as the stock of data in Common Crawl, at a cost of hundreds of billions of dollars in wages.}

Out of the remaining strategies for circumventing public human text data bottlenecks, we identify three broad categories of techniques that appear particularly promising. These are: a) using models themselves to generate more data, b) multimodality and transfer learning, which involves training language models on other existing datasets (e.g. from different domains), and c) using non-public data.

\subsection{AI-generated data}

OpenAI alone reportedly generates 100B words per day \cite{griffin2004altman}. Within a year, this corresponds to around 36.5T words, not far from our estimates of the total number of high-quality words in Common Crawl. If outputs are accumulated across different models and across time, the growth in the stock of training data could expand dramatically in principle, assuming this approach works.

However, the evidence for the effectiveness of training on generated (synthetic) data is currently mixed. One challenge is that models might lose information about the original human data distribution, such that iteratively training on model outputs results in increasingly homogeneous and unrealistic outputs \cite{shumailov2023curse}. More generally, repeatedly training on synthetic data can yield diminishing or even negative returns \cite{singh2023human}, and worse scaling behavior \cite{fan2023scaling, dohmatob2024tale}. These challenges can be mitigated to some extent by using training data with greater diversity \cite{fan2023scaling, openai2019solving}, or by training on a mixture of human-generated and synthetic data \cite{gunasekar2023textbooks, shumailov2023curse, gerstgrasser2024model, alemohammad2023selfconsuming}. 

On the other hand, training on synthetic data has shown much promise in domains where model outputs are relatively easy to verify, such as mathematics, programming, and games \cite{yang2023leandojo, liu2023tinygsm, haluptzok2023language}.\footnote{Verification processes can be used as training signals which guide the data generation and improve performance \cite{zhang2023chainofthought, huang2022large}.} For example, AlphaZero \cite{silver2017mastering} was famously trained using self-play, and more recently AlphaGeometry \cite{Trinh2024} was trained purely using synthetic data from attempts to solve geometry problems. What is less clear is whether the usefulness of synthetic data will generalize to domains where output verification is more challenging, such as natural language.\footnote{Despite this, researchers have attempted to train models on synthetic feedback, such as using model-generated critiques to prevent certain behaviors \cite{bai2022constitutional, burns2023weaktostrong, irving2018ai}. These approaches highlight potential advantages of synthetic data, including avoiding difficulties in generating human feedback at scale \cite{burns2023weaktostrong, khan2024debating, saunders2022selfcritiquing, michael2023debate}.}

We consider synthetic data to be one of the most promising avenues for circumventing data bottlenecks because of its potential to produce training data at an massive scale, its demonstrated success in certain domains, and the existence of potential strategies to mitigate the challenges associated with its use.

\subsection{Multimodal and transfer learning}

Another option is to go beyond \emph{text} data, and train models on data from other domains or non-text modalities, like images. Appendix \ref{appendix:nontext-data} includes some rough estimates of the stock of data for some of the most prominent modalities, concluding that current video and image stocks are not large enough to prevent a data bottleneck.

But there are other sources that can provide orders of magnitude more data of various types (e.g. financial market data, scientific databases, etc.). For illustration, \cite{Stephens2015BigDA} forecasts growth rates of between 2-40 million terabytes of genomics data every year by 2025. 

While it is not clear that leveraging data-rich domains for language modeling is always possible, there is already evidence that this is feasible in some specific cases. For instance, current frontier models like GPT-4V are trained on both image and text data \cite{2023GPT4VisionSC, gemini}. \citet{aghajanyan2023scaling} study this question for several modalities of data and show that these modalities have some synergy with text, when training on an even mix of both. In general, better understanding the feasibility of transfer learning would require further research, such as scaling laws for transfer learning \cite{hernandez2021scaling}.

\subsection{Using non-public data}

While the indexed web is vast, its size is small relative to the \emph{deep web}: the part of the web that is not accessible by search engines. The largest components of the deep web are closed content platforms like Facebook, Instagram or Twitter. While part of these platforms are indexed, the vast majority is not. Another large reservoir of non-public text data can be found in instant-messaging applications like WhatsApp or Facebook Messenger.

In Appendix \ref{appendix:nonpublic-data}, we estimate that content platforms and instant messaging apps both contain on the order of one quadrillion tokens. Combining this with the similarly-sized upper estimate of the raw stock of text in the indexed web, the total stock could reach 3 quadrillion tokens. This increase would delay a data bottleneck by about a year and a half relative to using only data from the indexed web.

However, the non-public stock seems unlikely to be as useful as indicated by our estimate. First of all, training on this data would be a grave violation of the privacy of the users who submitted the data to platforms without expecting it to be used for training AI models and probably would face legal challenges. Second, the quality of social media content is probably substantially lower than that of web content. Finally, this data is fragmented across several closed platforms that are controlled by different actors, so it is unlikely that all of it can be used in a single training run.

\subsection{Data efficiency techniques}

According to \citet{ho2024algorithmic}, training techniques and algorithms for LLMs have been improving at a rate of 0.4 OOM/y [95\%: 0.1, 0.8], meaning that roughly 0.4 fewer OOMs of compute are needed each year to achieve the same levels of performance. This is partially due to more efficient data use. Similarly large gains in sample efficiency has been found for reinforcement learning \cite{dorner2021measuring}. Although we do not know precisely what fraction of LLM efficiency gains result from ``doing more with less data," it is possible that improvements in data efficiency are occurring at a pace that could compensate for the exhaustion of data stocks. %As a result, even if data scaling were to halt, it might not necessarily lead to a dramatic slowdown in the progress of AI performance.

\subsection{Other techniques}

Another possibility is learning from interactions with the real world, which might include LLMs training on the messages received from users or, if ML models become sophisticated enough to act autonomously, learning from sensory observations or from the results of real-world experiments. This form of learning will probably become necessary at some point if AI models are to surpass human knowledge about the real world.

One additional broad category of techniques is data selection, in which we include techniques like pruning \cite{marion2023more}, domain composition tuning \cite{xie2023doremi}, and curriculum learning \cite{campos2021curriculum}. However, we do not find this class of techniques very promising since the gains tend to be modest.\footnote{See, for example, \citet{marion2023more}, where pruning provides less of a benefit than scaling the dataset by 3x, or \citet{tirumala2023d4}, where the benefit of selection is similar to that of scaling datasets by 20\%. In general, the benefit of these techniques is likely limited by the fraction of performance-degrading datapoints in the dataset.}
\section{Discussion}
\label{sec:discussion}

In this paper, we examine the challenges and opportunities that lie ahead for scaling machine learning systems, particularly in light of the finite nature of public human text data. Our analysis reveals a critical juncture approaching by the end of this decade, where the current reliance on public human text data for training ML models may become unsustainable. Despite this looming bottleneck, we identify transfer learning and self-generated data as viable and promising pathways that could enable the continued growth and evolution of ML systems beyond the constraints of public human text data.

Our conclusions are thus twofold. On the one hand, we expect that the current paradigm based on public human text data will not be able to continue a decade from now. On the other hand, it is likely that alternative sources of data will likely be adopted before then, allowing ML systems to continue scaling.

While our arguments about alternative sources of data are mostly qualitative, a better understanding of data quality could make it possible to make quantitative estimates of the benefits of transfer learning and synthetic data. For example, scaling experiments for transfer learning could be used to quantify the proximity or synergy between different distributions \cite{hernandez2021scaling, aghajanyan2023scaling} and identify new datasets which can effectively expand the stock of data.

This paper does not explore certain considerations that might be relevant for understanding the future role of data. Firstly, the choice of data should depend on the desired skills or capabilities of the model. Identifying economically or scientifically valuable skills and the datasets needed to teach them could reveal critical data gaps. Secondly, future ML breakthroughs, such as systems capable of autonomous real-world exploration and experimentation, might change the dominant source of information for learning.

\section{Conclusion}

We have projected the growth trends in both the training dataset sizes used for state-of-the-art language models and the total stock of available human-generated public text data. Our analysis suggests that, if rapid growth in dataset sizes continues, models will utilize the full supply of public human text data at some point between 2026 and 2032, or one or two years earlier if frontier models are overtrained. At this point, the availability of public human text data may become a limiting factor in further scaling of language models.

However, after accounting for steady improvements in data efficiency and the promise of techniques like transfer learning and synthetic data generation, it is likely that we will be able to overcome this bottleneck in the availability of public human text data. 

It is important to acknowledge the inherent uncertainty in making long-term projections, especially considering the rapid pace of advancements in the field of AI. Our results highlight the need for further research to quantify data efficiency growth rates and the potential performance gains from emerging methods. Additionally, future work should explore the feasibility and effectiveness of transfer learning from diverse data domains and the impact of synthetic data generation on model performance, among other things.

\newpage

\section*{Acknowledgments}

We thank the ICML reviewers, Nuño Sempere, Eli Lifland, Ege Erdil, Matthew Barnett and Joshua You for their thoughtful comments and contributions to this paper.

\section*{Impact Statement}

The practice of scraping data from the web and using it for large-scale training of AI systems raises important issues regarding fairness and justice. In particular, there are strong arguments in favor of compensating the creators of the data used to train these systems. While AI has the potential to greatly increase productivity and overall welfare, it is important to factor in these justice-related considerations to ensure that the benefits are distributed equitably.

Our work suggests that data from social media platforms and messaging apps could serve as a significant and valuable resource for training AI systems. However, using this type of data for training raises serious privacy and security concerns. Without proper safeguards in place, sensitive personal information from these platforms could be exposed to users of the AI systems .The risks associated with using non-indexed platform data for training may be substantial enough to outweigh the potential benefits gained from using this data.

\nocite{*}
\bibliography{references}
\bibliographystyle{icml2024}

\appendix

\section{Theoretical growth model of the web}
\label{appendix:reddit}

We explain in more detail our theoretical model of data accumulation rates developed in Section \ref{sec:internet_pop} and check it on Reddit submission data. The model is explained in Equation \ref{eq:stock-internet_users}.

\begin{figure}[ht]
  \small
  \centering
  \includegraphics[width=0.45\textwidth]{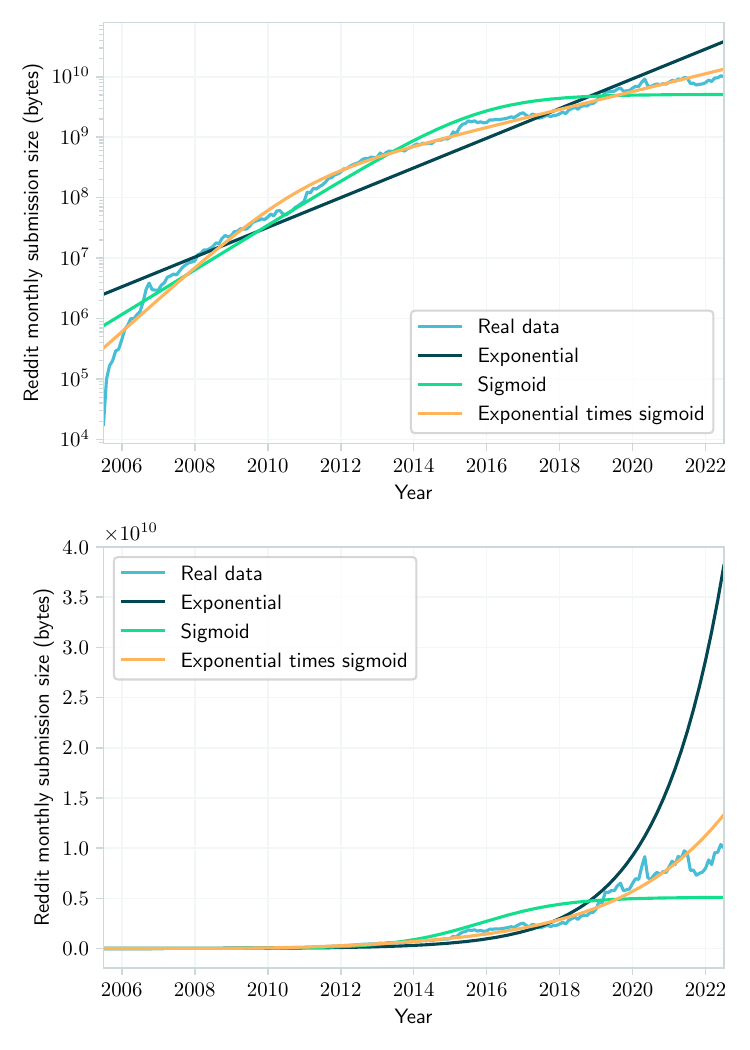}
  \caption{\small Monthly user submissions to Reddit, in linear scale(down) and log scale(up). While the three functions appear to fit the data reasonably well in the log scale, the linear plot shows that the sigmoid times exponential function predicts much better the recent years.}
  \label{fig:reddit_test}
\end{figure}

A purely exponential model cannot reproduce the decrease in the growth rate of Reddit submissions over time, while a purely sigmoidal model plateaus at zero growth. The exponential times sigmoid model is able to better capture the deceleration in submission size growth (see Figure \ref{fig:reddit_test}).

In our actual model, the slowdown in population growth (which becomes subexponential) leads to additional deceleration, but the time period covered by the Reddit submission dataset seems too short for slowing population growth to be noticeable in the data.

\begin{tcolorbox}[title=Projection Based on the Number of Internet Users ] 
\begin{equation}
    S_{IU}(y) = D_{y_0}\int_{1950}^y \frac{H(x) \sigma\left((x-s_0)\times0.15\right)}{H(y_0) \sigma\left((y_0-s_0)\times0.15\right)}dx
    \label{eq:stock-internet_users}
\end{equation}

where \(D_{Y_0}\) is the amount of data produced in some reference year \(Y_0\), \(H(Y)\) is the projected human population in a certain year, and the sigmoid \(\sigma\) models internet penetration, which is approximately 0\% in 1950 (this why we choose it as the initial point for the integral) and 50\% in $s_0=2016$. 0.15 is a fitted scale parameter. The integral represents the total number of person-years of internet use, normalized by the internet use in the reference year.
\end{tcolorbox}

\section{Estimating the size of the indexed web}
\label{appendix:google-index}

The ``indexed web'' comprises those web pages that are included in the indices of search engines. In particular, since Google is the most popular search engine worldwide, we tried to estimate the number of web pages in Google's index.

We replicate the methodology of \citet{vandenBosch2016}. We calculate the frequency of words in a large corpus of clean web documents: the RefinedWeb dataset \cite{penedo2023refinedweb}. Then we select a set of words at logarithmically equidistant intervals of frequency, called ``pivot words'' in \citet{vandenBosch2016}. Using the number of results that Google reports when searching each of the pivot words, we can extrapolate the total size of Google's index, assuming that the frequencies of the words are similar in our corpus and in Google's index.

\begin{figure}[ht]
  \small
  \centering
  \includegraphics[width=0.45\textwidth]{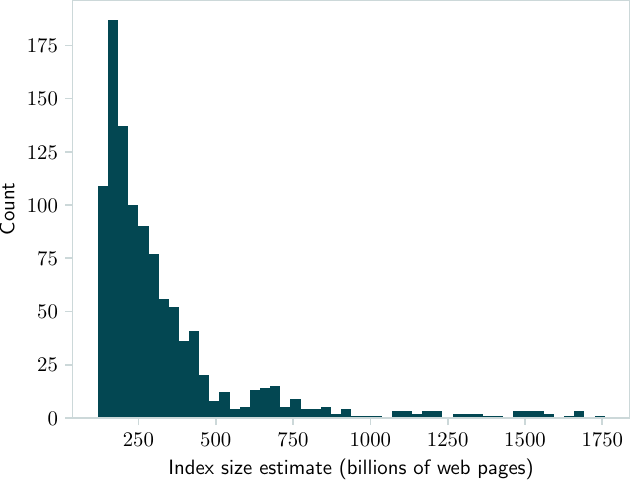}
  \caption{\small Histogram of the estimates of the size of Google's index from each pivot word.}
  \label{fig:index_size}
\end{figure}

Each pivot word provides a noisy estimate of the total size, so we take the average to arrive at a more robust estimate. Using 100 pivot words, the distribution of estimated sizes is approximately log-normal, with a mean of 330B web pages, a median of 250B web pages, and a 95\% CI between 100B and 1200B (see \ref{fig:index_size}). This is about 4 times more than Common Crawl, which only has 75B unique urls.

Our results are substantially higher than those obtained by \citet{vandenBosch2016}. This is mostly because we retrieved the number of Google results for each word using the Google Custom Search JSON API. However, \citet{vandenBosch2016} used the numbers shown in Google's web interface, which are around half as big for the same search terms.

We evaluated whether a change in the relative frequency of words in the web over time might lead to inaccuracies when the same frequencies are used to estimate the size of the index across several years. To do this, we computed word frequencies for webs sampled in different years between 2013 and 2021. The difference in the resulting estimates was smaller than 10\%, so we conclude that this is unlikely to add significant noise to the estimate.

\subsection{Distribution across languages}

The exact distribution of languages in the web is hard to estimate. In Common Crawl, around 45\% of webpages is in English. This is broadly consistent with the 58.8\% reported by the ``Digital 2023: Global Overview Report'' \cite{datareportalDigital2023}. However, the Observatory of Linguistic and Cultural Diversity in the Internet reports that in 2023 around 20\% of the content of the web was in English \cite{obdilciIndicatorsPresence}.

If Common Crawl does not constitute a representative sample of the web, then our previous method for estimating the size of the web might be biased. However, a reduction in the English share from 45\% to 20\% would only increase the estimate of the total size by a factor of 2.25, which is not enough to significantly change our conclusions.

\subsection{Growth in the size of the indexed web}

As documented in \citet{vandenBosch2016}, the size of Google's index has varied significantly over the past decade. However, this variation does not seem to follow any monotonic temporal pattern, and instead consists of seemingly random movement around a stable mean. The fact that the temporal mean of the index size has not increased over time is surprising: the number of internet users grows by about 1.8\% a year \cite{datareportalDigital2023}, the number of pages in Wikipedia grows by 2.6\% per year \cite{wikipediaSize}, and the global size of IP traffic grew by 24\% per year in 2016 \cite{ciscoIndex}. Given these facts, it seems unlikely that the overall number of web pages has not grown significantly.

We propose several hypotheses. One is based on link rot, the phenomenon of web links becoming inaccessible over time due to deletion, failure, restructuring of web sites, or other reasons. There have been several estimates of the rate of link rot over time, but most claim that between 2\% and 16\% of links break during a year \cite{Howell2022TheCP, Loan2020TheDA, Loan2023GivingLT, ott2022reference, satyanarayana2022web, Zittrain2021ThePO}. Some of the broken links will be replaced by new links to the same content (think of a change in domain name, for example, in which the new domain will point to the same documents as the old one), but part of that content will be irreversibly lost. If the rate of growth of the web is similar to the rate of link rot, the two effects might partially cancel each other and lead to a more stable index size.

Another possibility is that Google is keeping the size of their index within a fixed range due to economic or engineering considerations. In this case, perhaps the web pages deemed least valuable or useful are eliminated from the index as more valuable pages appear. It is known that Google does not index all the pages they crawl \cite{googleblogKnewBig} due to quality considerations.

A final possibility is that all of our estimation methods are biased towards content in the Anglosphere or the West more generally. It is possible that both Google and Common Crawl are not representative examples of the global web, but only a part of it. Since most of the recent growth in the number of internet users has been in non-Western countries \cite{datareportalDigital2023}, it is possible that our estimates are missing this growth.

\section{Non-public text data}
\label{appendix:nonpublic-data}

The term ``deep web'' refers to that portion of the web that is not accessible by search engines. While this includes many categories of data, in this section we focus on closed\footnote{By closed, we mean that they usually require credentials to access and limit the visibility of their content to unregistered users.} content platforms, since they attract a large share of all web traffic.\footnote{According to Similarweb, Google, YouTube, Facebook, Instagram, Twitter and Baidu account for a third of global web traffic \cite{similarweb}.}

Given the power-law nature of web pages usage, we can obtain a fairly reliable estimate of the total size of the deep web just by examining a few of the most-visited platforms. In particular, we estimate the amount of data in Facebook, Instagram and Twitter, three of the largest social media platforms, as well as Reddit, an indexed and open social media platform that we use as a sanity check. We then divide the estimate of the stock of each platform by its share of global traffic to arrive at a global estimate.

\paragraph{Facebook:} Facebook has about 3B users, who on average make one post and five comments per month \cite{datareportalFacebookUsers}. From a sample of 70,000 posts and 200,000 comments \cite{krebs2017social} we calculate the average post has 60 tokens, while the average comment has 26 tokens.\footnote{This particular sample might not be representative of the all the content in Facebook, but we think it's unlikely to be too different from the true averages.}. This corresponds to 10T tokens being produced each year. Since Facebook has had a roughly similar number of users for the past 10 years or so, its total stock is around 100T tokens.

\paragraph{Instagram:} Instagram has 1.6B users \cite{datareportalDigital2023}, who in aggregate make about 66,000 posts per minute \cite{domoDataNeverSleeps}, or about 100 million per year. The average post has about 8-16 tokens \cite{thorgrenTemporalDO}, for a total of around 800B tokens generated per year. Over the past 10 years or so, this corresponds to 8T tokens.

\paragraph{Twitter:} Twitter has close to 300M monthly active users. Each user posts 1.3 times per day, and the average post has about 7 tokens \cite{gdeltprojectVisualizingTwitterx2019s}. This corresponds to about 1.5T tokens per year, and since Twitter has had a roughly constant number of users for the past 12 years, the total stock is around 17T tokens.

\paragraph{Reddit:} Reddit is public and indexed, so accurate statistics are available. The number of posts per day was about 60,000 in 2020, while the number of comments was about 500,000 \cite{baumgartner2020pushshift}. These have grown close to linearly since 2011, so the total is equivalent to about 7 years of submissions at that rate. Each post has on average 48 tokens, and each comment has 21.\footnote{These are estimated from 3M posts and 4.5M comments in the Pushshift dataset.} This corresponds to a total of 75B tokens per year, or about 600B tokens in total.

Finally, we divide each estimate by the share of traffic of the corresponding platform to arrive at an estimate of the total stock in the web. We take the average of the estimates of each platform. As shown in Table \ref{table:deep-web}, the results indicate that the size of the deep web is roughly comparable to the size of the indexed web, so using this source of data would only delay a data bottleneck by a couple years.

\begin{table}[h]
\centering
\small
\begin{tabularx}{\columnwidth}{@{}llll@{}} % Corrected to four columns; two regular right-aligned
\toprule
Platform & Tokens & Traffic share & Total stock estimate \\ \midrule
Facebook    & 100T         & 3.6\%         &    3000T   \\
Instagram   & 8T          & 1.5\%         &    580T    \\
Twitter     & 17T         & 1.3\%         &    1370T    \\
Reddit      & 600B        & 0.44\%        &    140T     \\
Geom. mean   &             &               &    760T    \\
\bottomrule
\end{tabularx}
\caption{Estimates of the stock of data in the deep web. Estimates of the total stock are produced by dividing the estimate of the number of tokens by the share of traffic. These estimates are rough, and expected to be accurate only up to an order of magnitude.}
\label{table:deep-web}
\end{table}

\subsection{Instant messaging and email}

Instant messaging applications are widely used: Facebook Messenger alone has over a billion users \cite{datareportalFacebookMessenger}. They consequently contain a vast amount of non-indexed data. In 2020, the messaging apps owned by Meta (then Facebook) were processing 100B messages per day \cite{fbMessengerIntroducing}. In a sample of 6 million WhatsApp messages, \citet{Rosenfeld_whatsapp_22} found that the average message had 5.66 words, or about 4.5 tokens. This corresponds to 165T tokens generated per year, for a total stock over a quadrillion tokens.

Note that the quality of instant messaging data relative to web data is not well documented, so this source of data might be much less useful for training than our calculation of the number of tokens suggests. That said, even if we take the number of tokens at face value, it is comparable to the stock of text in the indexed and deep web, and therefore instant messaging data only increases the stock of text by 50\%, which would delay a data bottleneck by less than one year.

Emails could also be a large source of data, comparable to the overall size of the deep web. There are reportedly over 300B emails received per day \cite{radicati}, which amounts to around 100T emails per year. It is unclear how many tokens of unique content this contains. It is likely that most are "mass emails" such as advertisements or newsletters such that most received emails are not unique or have substantial duplicate content. If we conservatively suppose that 10\% of emails are unique and contain about 50 words on average (consistent with some existing estimates, see \citet{taniguchi-etal-2020-large}), this would represent about 625T tokens, which is the same order of magnitude as our estimates of the stock of deep web text.

\section{Non-text data}
\label{appendix:nontext-data}

There are significant sources of public data in modalities different than text. Most notably, the web contains large quantities of images and videos which might be used to train multimodal systems. However, we believe that the amount of useful information contained in these additional sources of data is not enough to meaningfully change our conclusions. In this appendix we arrive at some rough estimates for the amount of data available for other common modalities to justify this conclusion.

%Many outcomes from social dynamics have heavy-tailed distributions, and in particular this is the case for the success of web content platforms like social media sites: the number of visits to web platforms follows a power-law distribution, so most of the content on the internet is concentrated in the top most-used platforms. For this reason, we focus on estimating the amount of content in these top platforms.

\paragraph{Images} It is hard to know exactly how many pictures are taken globally per year, but a reasonable estimate is on the  order of a couple trillion \cite{riseabove2021}. \footnote{This is consistent with the average human taking about 100 pictures per year.} \citet{henighan2020scaling} estimates that one image has at least as much information as around three tokens of text. Meanwhile, image encoders often use hundreds of tokens per image \cite{dosovitskiy2021image}, but there is probably significant redundancy in this representation. We take the geometric average of these two extremes to arrive at a reasonable middle point of $\sim$30 tokens per image. Assuming that this rate of image capture has been maintained for 10 years, this corresponds to a few hundred trillion tokens, roughly the same scale as the raw size of Common Crawl. For this reason, including images in our model is not likely to produce large changes in our results.

\paragraph{Video} In the case of video, YouTube is the most-used video hosting platform worldwide. More than 500 hours of video are uploaded to YouTube every minute \cite{blogYouTubePress}, which corresponds to 1 trillion seconds of video uploaded per year. Using again the hypothesis that the information in an image corresponds to around 30 tokens of text,\footnote{It is possible that the useful information in videos is overwhelmingly in the spoken words, rather than the images. Since humans speak at 100-160 words per minute, the amount of tokens per second of audio cannot be higher than 4-5, so 30 tokens per second of video is still an upper bound.} assuming each second of video is as valuable as an independent image, and maintaining the same rate of video upload for 10 years, we arrive at an estimate of 100 trillion text-token-equivalents. YouTube represents 7\% of the share of internet traffic \cite{similarweb}, so if we assume that the share of content is similar to the share of traffic,\footnote{While this might not be exactly true, it seems hard to argue that other platforms contain a larger quantity of varied and useful videos.} the total stock of video might be on the order of one quadrillion tokens. Since this is similar to the stock of data in the web, including video data in our estimates of the stock would not significantly change our results.

One relevant consideration regarding video and images is that their production is much easier to scale than text. 7B CMOS image sensors were produced in 2020 \cite{cmosImageSensor}. If all of them were used for recording, in a single year they could produce 2e17 seconds of video, 1000 times more than the current stock in the web.

\paragraph{Exotic modalities} \citet{Stephens2015BigDA} estimated that astronomy and genomics produce several million terabytes of compressed data per year. At face value, this corresponds to a stock of roughly 1e18 tokens, a thousand times larger than our estimates of the stock for text, images and video.

However, these modalities of data have extremely high redundancy and significant noise. The amount of synergy that these modalities have with text is also an open question. For this reason, we currently cannot evaluate whether these alternative modalities can provide a lasting source of data.

\section{Tokenization Schemes Across Datasets}
\label{appendix:tokenization}

This appendix examines the performance of various tokenization schemes across different datasets by analyzing the average number of tokens produced and the average number of characters per token. The data presented in Tables \ref{table:refinedweb}, \ref{table:gpt2-datasets}, and \ref{table:various-tokenizers} collectively demonstrate that common tokenizers produce a similar number of tokens across a diverse range of texts.

\subsection{Tokenization in RefinedWeb Dataset}
Table \ref{table:refinedweb} compares three tokenizers (BERT, GPT2, and XLMNet) on 1000 random web pages in the first segment of the RefinedWeb dataset, showing that the average number of tokens produced by each tokenizer is similar, ranging from 621 to 653, with a consistent number of characters per token (around 4.2 to 4.4). In this dataset the average length of a word in bytes including whitespace is 5.5, so on average there are 4.4/5.5 = 0.8 words per token.

\begin{table}[h]
\centering
\begin{tabularx}{\columnwidth}{@{}lXrr@{}} % Corrected to four columns; two regular right-aligned
\toprule
Tokenizer & Mean Tokens & Chars/Token \\ \midrule
BERT      & 621         & 4.4         \\
GPT2      & 645         & 4.2         \\
XLMNet    & 653         & 4.2         \\
\bottomrule
\end{tabularx}

\caption{Tokenization schemes in the first segment of RefinedWeb}
\label{table:refinedweb}
\end{table}
\subsection{Tokenization Using GPT2 Tokenizer Across Various Datasets}
Table \ref{table:gpt2-datasets} demonstrates the performance of the GPT2 tokenizer across various datasets, with the characters per token metric remaining relatively consistent, ranging from 2.22 to 4.15. This suggests that the GPT2 tokenizer produces a similar number of tokens per character across diverse text data, despite differences in the nature and average character length of the datasets.

\begin{table}[h]
\centering
\begin{tabularx}{\columnwidth}{@{}lXrr@{}} % Corrected to four columns; two regular right-aligned
\toprule
Dataset       & Texts & Avg Char Length & Chars/Token \\ \midrule
Enron Emails  & 1010  & 1618            & 3.44        \\
FreeLaw       & 5094  & 15707           & 3.53        \\
GitHub        & 18337 & 5238            & 2.53        \\
EuroParl      & 133   & 62975           & 2.5         \\
DM-Mathematics& 2007  & 8194            & 2.22        \\
ArXiv         & 2434  & 47345           & 3.05        \\
Books3        & 301   & 587352          & 4.15        \\
\bottomrule
\end{tabularx}
\caption{GPT2 Tokenizer performance across multiple datasets}
\label{table:gpt2-datasets}
\end{table}

\subsection{Performance of Modern Tokenizers on Selected Datasets}
Table \ref{table:various-tokenizers} compares the performance of multiple tokenizers used in modern models (Mixtral-7B, Command-plus-R, and cl100k\_base/GPT-4) on selected datasets. The characters per token metric remains fairly consistent for each tokenizer across the datasets and always between 2 and 5, indicating that the choice of tokenizer does not significantly affect the number of tokens produced per character.

\begin{table}[h]
\centering
\begin{tabularx}{\columnwidth}{@{}lXrr@{}} % Corrected to four columns; two regular right-aligned
\toprule
Tokenizer       & Dataset             & Chars/Token \\ \midrule
Mixtral-7B      & Books3              & 3.8         \\
                & GitHub              & 2.88        \\
                & Chinese Modern Poetry & 2.17      \\
Command-plus-R  & Books3              & 4.17        \\
                & GitHub              & 3.29        \\
                & Chinese Modern Poetry & 3.22      \\
cl100k\_base    & Books3              & 4.31        \\
                & GitHub              & 3.78        \\
                & RefinedWeb          & 4.41        \\
                & DM-mathematics      & 2.25        \\
                & Chinese Modern Poetry & 2.1       \\
\bottomrule
\end{tabularx}
\caption{Performance of various tokenizers across selected datasets}
\label{table:various-tokenizers}
\end{table}

The consistency in the characters per token metric across different tokenizers and datasets supports the conclusion that the conversion from the number of words to the number of tokens is roughly independent of the tokenizer used. This finding has implications for the comparability of tokenization results across studies using different tokenizers and datasets. Further statistical analysis, such as examining the variance or standard deviation of the characters per token metric, could provide additional insights into the consistency of tokenization across datasets and tokenizers.

\section{Overtraining in the context of data scarcity}
\label{appendix:overtraining_model}

In this appendix we sketch a model of optimal scaling decisions under data scarcity. In particular, we examine how the decision to overtrain models might be affected by data scarcity.

Our starting point is the parametric scaling law of \citet{hoffmann2022}, which predicts the reducible loss of a model $L$ given its number of parameters $N$ and the size of its training dataset $D$ (see Equation \ref{eq:red_loss}). \citet{hoffmann2022} derive from this scaling law a relation between the sizes of the model and the dataset that minimize the reducible loss of their model given a fixed training compute budget. In particular, in compute-optimal models the ratio $D/N$ is around 20. We call this the Chinchilla scaling law, and we call models that follow it Chinchilla-optimal.

Models for which the ratio $D/N$ is above the Chinchilla-optimal ratio are commonly called overtrained, while models that are below that ratio are called undertrained. At a fixed training compute budget, overtrained models require less compute during inference but more data during training. This is currently attractive for developers, as compute is relatively scarce compared to data. As a consequence, some well-known models, like Llama 3 \cite{llama3}, are overtrained.\footnote{Llama 3 has 70B parameters and was trained on 15T tokens, so it has 214 tokens per parameter, 11x more than the Chinchilla-optimal ratio.}

\begin{tcolorbox}[title=Profit maximization problem] 
\begin{align}
    \text{maximize} \quad & I (P - 2N) - 6ND \label{eq:profit} \\
    \text{s.t.} \quad & 6ND + 2NI = C_0 \label{eq:comp_cost} \\
    \text{where} \quad & I = I_0 (AN^{-a} + BD^{-b})^{-r} P^{-h}
    \label{eq:inf_demand}
\end{align}

Here $N$ is the number of parameters of the model, $D$ is the size of the training dataset in tokens and $P$ is the price of each inference token in (some multiple of) dollars. $C_0$ is the total computational budget for training and inference and $I$ is the number of tokens produced during inference. $A, a, B$ and $b$ are fitted parameters of the scaling law, and $I_0, r$ and $h$ are parameters of the inference demand function. All parameters are positive.
\end{tcolorbox}

We now examine the relationship between overtraining and undertraining in the context of a data bottleneck. To simplify the analysis, here we ignore the cost of gathering data and focus on the computational cost of the model during training and inference. We assume that developers want to achieve the maximum possible profit within their computational budget, and that this computational budget includes both training and inference. In particular, given $N$ and $D$, as well as a certain number of inference tokens, $I$, the compute cost $C$ is given by Equation \ref{eq:comp_cost}.

We assume that inference demand is a function that increases with model quality and decreases with the price of inference. We use the inverse reducible loss $L^{-1}$ as a proxy for quality.\footnote{In general, the performance of a model is inversely correlated with the loss, and this relationship can be approximated by a power law \cite{henighan2020scaling}.} The functional form is given by Equation \ref{eq:inf_demand}.

The optimal scaling policy depends on the values of $r$ and $h$. If $h=0$ or $r=0$, demand for inference is independent of price or capabilities, respectively. If $h=1$, demand is constant in dollar terms. If $h>1$, demand is decreasing in dollar terms. Since demand decreases with dollar price for most goods, we try values of $h$ greater than one. Since the loss of a compute-optimal model scales as $\sim C^{-0.15}$, for the profit to increase as a function of training compute (which is what we would expect) $r$ must be higher than $0.15^{-1} \approx 7$.

\begin{figure}[ht]
  \small
  \centering
  \includegraphics[width=0.45\textwidth]{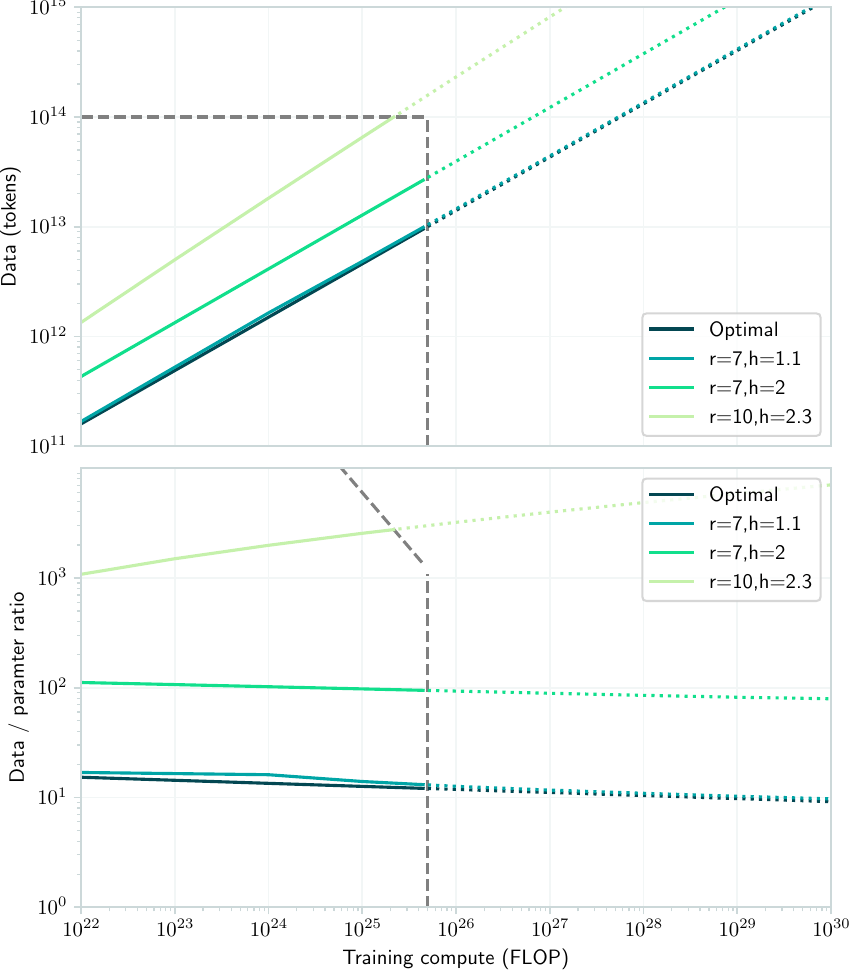}
  \caption{\small Scaling policies obtained by solving the optimization problem given by Equations \ref{eq:profit}-\ref{eq:inf_demand}. The dashed lines represent data and compute budgets of 100T tokens and 1e27 FLOP, respectively. Achievable configurations given these budgets are indicated by solid lines, while unachievable ones are indicated by dotted lines. Some scaling policies exhaust the compute budget first, meaning they are compute-constrained, while others exhaust the data budget first, indicating they are data-constrained.}
  \label{fig:opt_model}
\end{figure}

In general, higher values of $r$ and $h$ lead to greater returns to overtraining, due to additional demand and price sensitivity. Figure \ref{fig:opt_model} shows some optimal scaling curves for different values of these parameters of the inference demand function. In particular, the values $r=7, h=1.1$ lead to a policy that is very close to Chinchilla-optimal, while the values $r=7, h=2$ lead to about 5x overtraining, $r=10, h=2.4$ lead to a level of overtraining that increases with training compute, and is around 100x at 5e25 FLOP.

\begin{figure}[ht]
  \small
  \centering
  \includegraphics[width=0.45\textwidth]{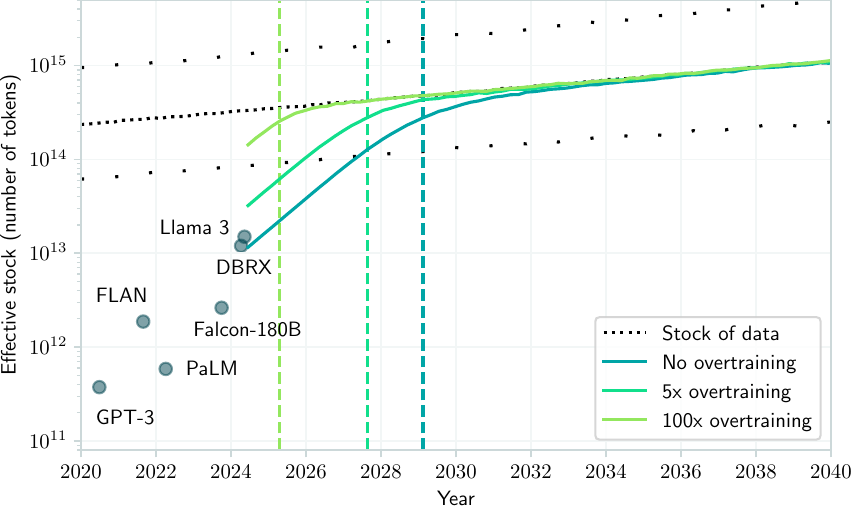}
  \caption{\small Compute-based projections of data usage and stock utilization years for the three profit-maximizing scaling policies from Figure \ref{fig:opt_model}. Only medians are shown for the dataset size projections.}
  \label{fig:overtr_projs}
\end{figure}

While Chinchilla-optimal scaling is bottlenecked by compute, under some assumptions the optimal scaling policy is instead bottlenecked by data. In any case, more overtraining leads to the stock of data being completely used earlier, as shown in Figure \ref{fig:overtr_projs}.

\begin{tcolorbox}[title=Scaling law for reducible loss] 
\begin{equation}
    L = AN^{-a} + BD^{-b}
    \label{eq:red_loss}
\end{equation}

Here $L$ is the reducible loss, $N$ is the number of parameters in the model, and $B$ is the size of the dataset in tokens. The values of the parameters $A, a, B$ and $b$ are taken from \citet{besiroglu2024chinchilla}.
\end{tcolorbox}

\section{Limits of undertraining in a data bottleneck}
\label{appendix:undertraining}

If data becomes scarce relative to compute, researchers might opt to undertrain increasingly large models on the existing stock of data. Using again the parametric scaling law found by \citet{hoffmann2022}, we can predict how much additional performance could be obtained from this approach.

\begin{figure}[ht]
  \small
  \centering
  \includegraphics[width=0.45\textwidth]{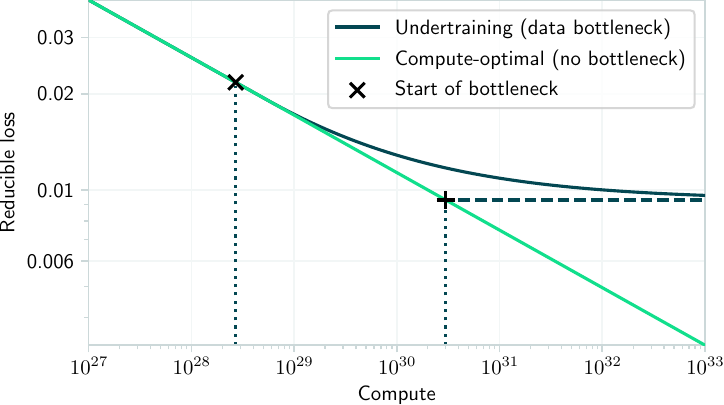}
  \caption{\small A toy model of undertraining in a data bottleneck scenario in which the stock of data is fixed at 300T tokens. The value of the loss is predicted using the parametric scaling law from \citet{hoffmann2022}, with the revised parameter estimates from \citet{besiroglu2024chinchilla}. The compute-optimal training compute corresponding to a dataset size of 300T tokens is shown in the left-most black vertical line. We also plot compute-optimal scaling with unlimited data for comparison.}
  \label{fig:toy_model_undertraining}
\end{figure}

In particular, we assume that the training data is fixed at 300T tokens and calculate the reducible loss predicted by Equation \ref{eq:red_loss} as we increase the training compute by adding more parameters to the model. Figure \ref{fig:toy_model_undertraining} shows the result from this model: undertraining can provide the equivalent of up to 2 additional orders of magnitude of compute-optimal scaling, but requires 2-3 orders of magnitude more compute. This is enough to sustain a decreasing rate of progress for 3-6 additional years before the final plateau.

\newpage

\end{document}